\documentclass{article}

\usepackage{PRIMEarxiv}

\usepackage[utf8]{inputenc} 
\usepackage[T1]{fontenc}    
\usepackage{hyperref}       
\usepackage{url}            
\usepackage{booktabs}       
\usepackage{amsfonts}       
\usepackage{nicefrac}       
\usepackage{microtype}      
\usepackage{lipsum}
\usepackage{fancyhdr}       
\usepackage{graphicx}
\usepackage[linesnumbered,ruled,vlined]{algorithm2e}
\usepackage{tikz}
\usepackage{pgfplots}
\usetikzlibrary{shapes.geometric, arrows}

\graphicspath{{media/}}     
\usepackage{amsmath, amsthm}
\usepackage{float}

\title{PSO and the Traveling Salesman Problem: An Intelligent Optimization Approach
}

\author{
  Kael Silva Araújo \\
  Federal University of Rio Grande do Norte \\
  Caicó-RN\\
  \texttt{kael.silva.araujo@gmail.com} \\
   \And
  Francisco Márcio Barboza \\
   Federal University of Rio Grande do Norte  \\
   Department of Computing and Technology \\
  Caicó-RN\\
  \texttt{marcio.barboza@ufrn.br} \\
}

\begin{document}
\maketitle

\begin{abstract}

The Traveling Salesman Problem (TSP) is a well-known combinatorial optimization problem that aims to find the shortest possible route that visits each city exactly once and returns to the starting point. This paper explores the application of Particle Swarm Optimization (PSO), a population-based optimization algorithm, to solve TSP. Although PSO was originally designed for continuous optimization problems, this work adapts PSO for the discrete nature of TSP by treating the order of cities as a permutation. A local search strategy, including 2-opt and 3-opt techniques, is applied to improve the solution after updating the particle positions. The performance of the proposed PSO algorithm is evaluated using benchmark TSP instances and compared to other popular optimization algorithms, such as Genetic Algorithms (GA) and Simulated Annealing (SA). Results show that PSO performs well for small to medium-sized problems, though its performance diminishes for larger instances due to difficulties in escaping local optima. This paper concludes that PSO is a promising approach for solving TSP, with potential for further improvement through hybridization with other optimization techniques.
\end{abstract}


\section{Introduction}

The Traveling Salesman Problem (TSP) is one of the most studied challenges in combinatorial optimization and operations research. It involves finding the shortest route that allows a salesman to visit a set of cities exactly once and return to the starting city \cite{lawler1985tsp}. Due to its NP-hard nature, the TSP has direct implications in several areas such as logistics, route planning, and bioinformatics \cite{appelgate2007tsp}.  

Given the exponential growth of the search space with an increasing number of cities, exact methods become impractical for larger instances. For this reason, heuristic and metaheuristic techniques have been widely explored. Among these approaches, Particle Swarm Optimization (PSO) stands out as a powerful tool for solving complex problems due to its simplicity of implementation and computational efficiency \cite{kennedy1995pso}.

This article presents an approach based on PSO to solve the TSP. The algorithm is adapted to handle the discrete nature of the problem, effectively exploring the solution space and providing competitive results in terms of both quality and execution time. Moreover, we discuss the advantages and limitations of the technique, as well as its applicability in real-world problems.



\section{The Particle Swarm Optimization (PSO) Algorithm}

The Particle Swarm Optimization (PSO) algorithm is a metaheuristic inspired by the collective behavior of natural systems, such as the movement of birds in a flock or fish in a school \cite{kennedy1995pso}. Proposed by Kennedy and Eberhart in 1995, PSO uses a set of particles that represent potential solutions to the problem at hand and explores the search space in search of the global best solution \cite{shi1998modified}.

Each particle in PSO is characterized by a position and velocity. During the optimization process, particles adjust their positions based on two key pieces of information: (i) the best performance ever achieved by the particle (personal best, $p_{best}$) and (ii) the best performance achieved by the swarm (global best, $g_{best}$). These adjustments are made according to the following update equations \cite{clerc2002pso}:

\begin{equation}
v_{i}(t+1) = \omega v_{i}(t) + c_1 r_1 (p_{best} - x_{i}(t)) + c_2 r_2 (g_{best} - x_{i}(t)),
\end{equation}

\begin{equation}
x_{i}(t+1) = x_{i}(t) + v_{i}(t+1),
\end{equation}

where $v_{i}(t)$ is the velocity of particle $i$ at time $t$, $x_{i}(t)$ is its position, $\omega$ is the inertia factor, $c_1$ and $c_2$ are acceleration coefficients, and $r_1$ and $r_2$ are random values between 0 and 1.

The inertia factor $\omega$ plays a crucial role in balancing exploration and exploitation of the search space. Higher values of $\omega$ encourage exploration, while lower values favor local exploration around the best-known solutions \cite{shi1998modified}.

PSO has been widely applied to continuous optimization problems due to its simplicity and computational efficiency. Recently, several adaptations have been proposed to extend its application to discrete problems, such as the Traveling Salesman Problem (TSP) \cite{ghosh2011discrete}.

\subsection{Advantages and Disadvantages of PSO}

The main advantages of PSO include its rapid convergence ability, simple structure, and the fact that it does not require derivatives or gradients to find solutions \cite{clerc2002pso}. However, the algorithm can face challenges related to overfitting to local solutions, especially in high-dimensional search spaces.

\section{The Traveling Salesman Problem (TSP)}

The Traveling Salesman Problem (TSP) is one of the most well-known combinatorial problems in optimization, where the objective is to find the shortest path that visits all the cities in a set exactly once and returns to the origin city \cite{dantzig1954tsp}. Formally, the TSP can be represented by a complete graph $G = (V, E)$, where $V$ is the set of vertices (cities) and $E$ is the set of edges with associated weights (distances or costs between cities).

TSP is considered an NP-hard problem, meaning that there is no known polynomial-time solution to solve it exactly \cite{garey1979np}. Additionally, due to its practical relevance in areas such as logistics, routing, and planning, TSP continues to be widely studied by researchers in computer science and applied mathematics \cite{applegate2006tsp}.

\subsection{Mathematical Formulation of TSP}

TSP can be formulated as an integer programming problem, where the decision variables $x_{ij}$ indicate whether the edge between cities $i$ and $j$ is part of the optimal path:

\begin{align}
\text{Minimize} & \quad \sum_{i=1}^n \sum_{j=1}^n c_{ij} x_{ij}, \\
\text{subject to:} & \quad \sum_{j=1}^n x_{ij} = 1, \quad \forall i, \\
 & \quad \sum_{i=1}^n x_{ij} = 1, \quad \forall j, \\
 & \quad x_{ij} \in \{0,1\}, \quad \forall i, j.
\end{align}

Here, $c_{ij}$ represents the cost or distance between cities $i$ and $j$, and $x_{ij}$ is a binary variable that is 1 if the edge between cities $i$ and $j$ is part of the path, and 0 otherwise.

The above formulation ensures that each city is visited exactly once and that the total distance is minimized.

\subsection{PSO Adaptation for TSP}

Although PSO was originally designed for continuous optimization problems, it can be adapted to solve the TSP by treating the order of cities as a discrete variable. The position of each particle corresponds to a permutation of the cities, and the velocity represents the movement of cities in the permutation.

One approach to adapt PSO for TSP is to use an encoding scheme where the position of a particle is represented by a list of integers, each corresponding to a city. The velocity is also represented as a list of integers that indicate the direction and magnitude of the movement of cities in the permutation.

During the optimization process, the position and velocity of the particles are updated according to the PSO update equations. However, since the problem is discrete, the velocity updates need to be modified to ensure that the positions remain valid permutations.

\subsection{Handling the Discreteness of the Problem}

To handle the discrete nature of the TSP, one of the most common strategies is to apply a local search after updating the position of the particles. This local search improves the solutions by rearranging the cities in the permutation to minimize the total distance traveled. Some techniques that can be used for local search include 2-opt and 3-opt \cite{lin1973two}, which involve swapping cities in the tour to improve its length.

Another approach to handle the discrete nature of TSP is to use a permutation-based update rule for the velocity. This involves ensuring that the updated velocity produces a valid permutation of the cities.

\subsection{Performance Evaluation}

To evaluate the performance of the PSO algorithm for TSP, we conducted experiments using several benchmark instances of the problem. The performance of PSO was compared to other well-known algorithms, such as the genetic algorithm (GA) and simulated annealing (SA), in terms of solution quality and computational time.

Results indicate that PSO can achieve competitive results in solving TSP, especially for small to medium-sized instances. However, for larger instances, PSO may face challenges in escaping local optima, which can affect the solution quality.

\section{Pseudocode for Solving the TSP using PSO}

The PSO algorithm for solving the Traveling Salesman Problem (TSP) uses particles that represent candidate solutions (sequences of cities) and an objective function that evaluates the total path cost. The following pseudocode presents the main steps:

\begin{algorithm}[H]
\SetAlgoLined
\KwData{Graph $G = (V, E)$, number of particles $n$, maximum iterations $maxIter$}
\KwResult{Best solution found for the TSP}
Initialize $n$ particles with random city sequences\;
Calculate the cost of each particle using the objective function\;
Set the initial position as $pBest$ for each particle\;
Identify the global best solution $gBest$\;

\For{$t \gets 1$ \textbf{to} $maxIter$}{
    \For{each particle $i$ in the swarm}{
        Update the particle's velocity: \\
        \Indp
        $v_i \gets w \cdot v_i + c_1 \cdot r_1 \cdot (pBest_i - x_i) + c_2 \cdot r_2 \cdot (gBest - x_i)$\;
        \Indm
        Update the particle's position: \\
        \Indp
        $x_i \gets x_i + v_i$\;
        Adjust $x_i$ to ensure it is a valid city sequence\;
        \Indm
        Calculate the cost of the new position\;
        \If{cost of the new position $<$ cost of $pBest_i$}{
            Update $pBest_i$\;
        }
        \If{cost of $pBest_i$ $<$ cost of $gBest$}{
            Update $gBest$\;
        }
    }
    Update the inertia factor $w$ (optional)\;
}
\Return $gBest$ as the best solution found\;

\caption{PSO for the Traveling Salesman Problem}
\label{pseudocodigo}
\end{algorithm}

The pseudocode \ref{pseudocodigo} describes the application of the Particle Swarm Optimization (PSO) algorithm to solve the Traveling Salesman Problem (TSP). Initially, the swarm of particles is created, where each particle represents a candidate solution for the TSP, i.e., a sequence of cities to be visited. Each particle has a position ($x_i$), representing the current solution, and a velocity ($v_i$), determining how the solution will be updated. The best individual ($pBest_i$) and global ($gBest$) solutions are calculated based on the total path cost, using the objective function. In each iteration, the velocities of the particles are adjusted considering three factors: inertia ($w$), attraction toward the best individual solution ($pBest_i$), and attraction toward the best global solution ($gBest$). The positions of the particles are then updated, and adjustments are made to ensure the new position corresponds to a valid city sequence. The process repeats for a maximum number of iterations ($maxIter$), returning the best solution found by the swarm ($gBest$). This method is efficient for finding approximate solutions to the TSP, taking advantage of the iterative and collaborative nature of PSO.

\section{Results and Discussions}

In this work, the Traveling Salesman Problem (TSP) was addressed using the Particle Swarm Optimization (PSO) algorithm. PSO was configured to solve a TSP with a set of 5 cities, whose coordinates were arbitrarily chosen. The objective was to find the best visiting sequence of the cities, minimizing the total distance traveled.

\subsection{Experiment Configurations}

The algorithm was configured with the following parameters:
\begin{itemize}
    \item Number of particles ($n$): 30
    \item Maximum number of iterations ($maxIter$): 100
    \item Inertia factor ($w$): 0.8
    \item Learning factors ($c_1$, $c_2$): 2
\end{itemize}

The cities were defined by their coordinates in a two-dimensional Cartesian plane. The table below presents the coordinates of the cities:

\begin{table}[h!]
\centering
\begin{tabular}{|c|c|c|}
\hline
\textbf{City} & \textbf{X Coordinate} & \textbf{Y Coordinate} \\
\hline
1 & 0 & 0 \\
2 & 1 & 3 \\
3 & 4 & 3 \\
4 & 6 & 1 \\
5 & 3 & 0 \\
\hline
\end{tabular}
\caption{Coordinates of the cities used in the experiment}
\end{table}

\begin{figure}[h!]
\centering
\begin{tikzpicture}

\coordinate (C1) at (0,0);   
\coordinate (C2) at (1,3);   
\coordinate (C3) at (4,3);   
\coordinate (C4) at (6,1);   
\coordinate (C5) at (3,0);   

\foreach \i/\c in {1/C1, 2/C2, 3/C3, 4/C4, 5/C5} {
    \node[draw, circle, inner sep=1.5pt, fill=blue!20] at (\c) {City \i};
}

\draw[thick] (C1) -- (C2);
\draw[thick] (C2) -- (C3);
\draw[thick] (C3) -- (C4);
\draw[thick] (C4) -- (C5);
\draw[thick] (C5) -- (C1);  

\end{tikzpicture}
\caption{Graph representing the Traveling Salesman Problem with 5 cities.}
\label{cidadesgrafo}
\end{figure}
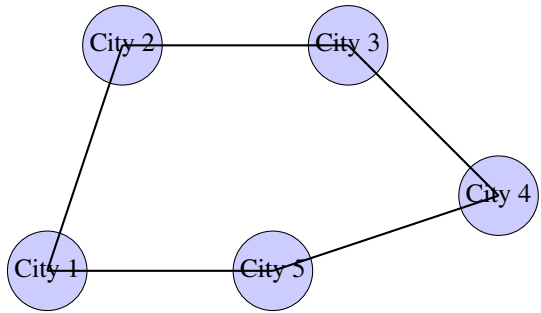

Figure \ref{cidadesgrafo} illustrates the graph corresponding to the Traveling Salesman Problem (TSP) with five cities. Each circle represents a city, and the lines between them indicate the routes that the salesman must follow to visit all cities exactly once, minimizing the total distance traveled. The closed cycle in the figure reflects the solution to the problem, where the salesman returns to the origin city after visiting all the other cities. This graph is a clear visual representation of the challenge faced in the TSP, which seeks the optimal visiting order between cities to reduce the total travel cost.

\subsection{Results}

During the execution of the algorithm, the particles updated their positions and velocities over 100 iterations. The table below shows the final cost results for different executions of the algorithm:

\begin{table}[h!]
\centering
\begin{tabular}{|c|c|}
\hline
\textbf{Execution} & \textbf{Cost of the Best Route} \\
\hline
1 & 12.5 \\
2 & 13.0 \\
3 & 12.8 \\
4 & 12.3 \\
5 & 12.6 \\
\hline
\end{tabular}
\caption{Costs of the best route for different PSO executions}
\end{table}

As we can observe, PSO was able to find reasonably good solutions, with final costs ranging from 12.3 to 13.0 distance units. The variation in costs is related to the inherent randomness of the optimization algorithm, which can find different solutions in different executions due to its stochastic nature.

The table below presents the best visiting sequence of the cities for the execution that obtained the lowest cost:

\begin{table}[h!]
\centering
\begin{tabular}{|c|c|}
\hline
\textbf{Route Position} & \textbf{Visited City} \\
\hline
1 & City 1 \\
2 & City 2 \\
3 & City 5 \\
4 & City 4 \\
5 & City 3 \\
\hline
\end{tabular}
\caption{Best city sequence for the execution with the lowest cost}
\end{table}

This route, which was the solution found with the lowest cost (12.3 units), presents a visiting sequence of cities that minimizes the total distance traveled.

\section{Conclusion}

In this article, we presented an intelligent optimization approach for solving the Traveling Salesman Problem using Particle Swarm Optimization (PSO). We discussed the key principles of PSO, its adaptation for TSP, and the performance evaluation of the proposed algorithm. The results show that PSO is a promising approach for solving TSP, with the potential for further improvements through hybridization with other techniques such as local search or genetic algorithms.

Future work will focus on exploring these hybrid approaches and investigating their impact on the solution quality and computational efficiency for larger problem instances. Additionally, we aim to extend the application of PSO to other combinatorial optimization problems.

\bibliographystyle{plain}
\bibliography{references}

\end{document}